\documentclass[letterpaper, 10 pt, conference]{ieeeconf}  % Comment this line out if you need a4paper

\IEEEoverridecommandlockouts                              % This command is only needed if 
\usepackage{comment}
\usepackage{graphicx}
\usepackage{siunitx}

\title{Open-Source Multi-Viewpoint Surgical Telerobotics}

\author{Guido Caccianiga$^{1}$, Yarden Sharon$^{1}$, Bernard Javot$^{1}$, Senya Polikovsky$^{2}$, Gökce Ergün$^{2}$,
        Ivan Capobianco$^{3}$,\\ André L. Mihaljevic$^{3}$, Anton Deguet$^{4}$, Katherine J. Kuchenbecker$^{1}$
\thanks{$^{1}$ Haptic Intelligence Department; 
$^{2}$ Optics and Sensing Laboratory, Max Planck Institute for Intelligent Systems, Stuttgart, Germany. 
$^{3}$ Department of General, Visceral and Transplant Surgery, University Hospital Tübingen, Tübingen, Germany.
$^{4}$ Laboratory for Computational Sensing and Robotics (LCSR) Johns Hopkins University, Baltimore, USA.}
\thanks{Corresponding author: Guido Caccianiga, \texttt{caccianiga@is.mpg.de}}
}

\begin{document}

\maketitle
\thispagestyle{empty}
\pagestyle{empty}

\section{Introduction}
Despite impressive development, current robots for minimally invasive surgery (MIS) are still at the beginning of their evolution. As these systems gradually become more accessible and modular, we believe there is a great opportunity to rethink and expand the \textit{visualization and control paradigms} that have characterized surgical teleoperation since its introduction several decades ago. While advances in computer vision and AI are increasingly applied in the surgical domain, their impact is often limited by the quality and quantity of the available clinical data. For example, the accuracy of spatiotemporal scene reconstruction and rendering is known to improve as the number of synchronous measurements taken from different perspectives increases. However, routine MIS and surgical robots feature only a single viewpoint (monocular or stereo). We conjecture that introducing a \textit{second adjustable observation angle in the abdominal cavity} would not only unlock novel visualization and collaboration strategies for surgeons but also substantially boost the robustness of machine perception toward shared autonomy.
    
Over the years, researchers have prototyped multi-viewpoint surgical imaging using trocars~\cite{Caccianiga2022DenseSurgery}, magnetic actuation~\cite{simi2013}, and directly grasping and positioning an additional stereo camera with a robotic tool~\cite{Avinash2019AStudy}. This last configuration, inspired by clinical practice with drop-in ultrasound probes, not only highlighted the advantage of dexterously controlling a second viewpoint, but also hinted at the possibility of teleoperating surgical tools from a different perspective. Our research expands these concepts by elaborating the clinical applications that could most benefit from multi-viewpoint robotic MIS (Section~\ref{Clinical}) and by creating and releasing high-performance technical solutions (Section~\ref{Technology}) that boost research and support future clinical validation (Section~\ref{Impact}).

\begin{figure}[t!]
    \centering
    \includegraphics[trim=0 132 0 0,clip,width=0.9\columnwidth]{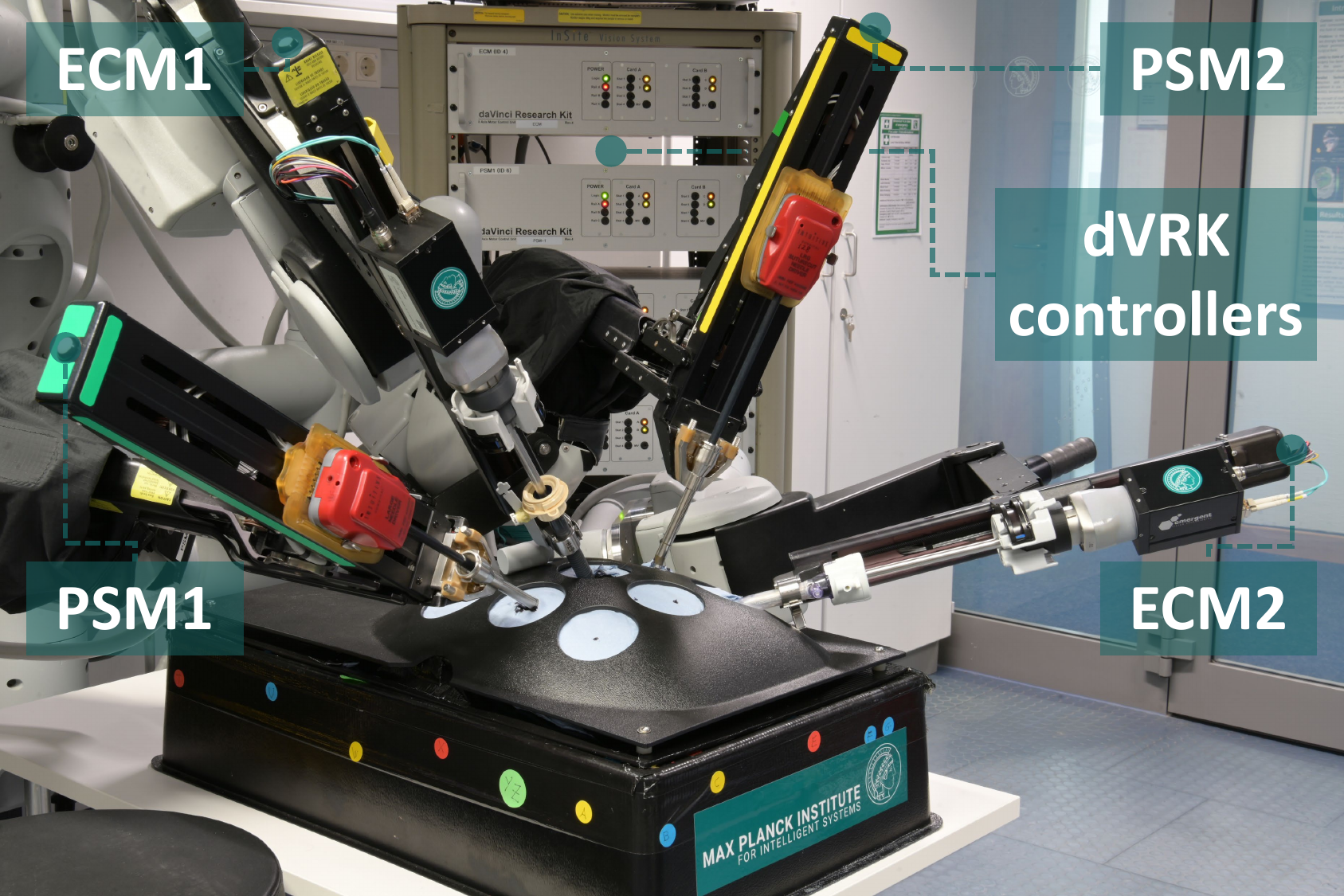}
    \vspace{-2mm}
    \caption{Our multi-viewpoint surgical robot (dVRK) featuring two standard instruments (PSM) and two camera manipulators (ECM); each ECM holds one of our synchronized, low-latency, stereo endoscope prototypes.}
    \label{Fig:dVRK}
    \vspace{-6mm}
\end{figure}

\begin{figure*}[t!]
    \centering
    \includegraphics[trim=0 0 0 0,clip,width=\textwidth]{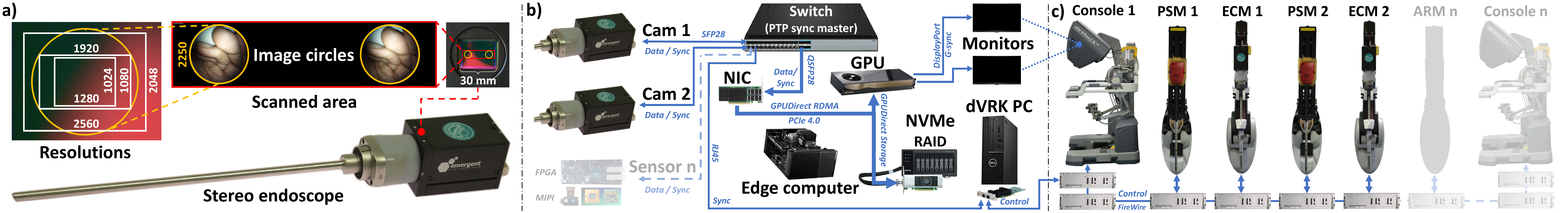}
    \vspace{-5mm}
    \caption{a) Our stereo endoscope prototype. Both images are captured by a single large global-shutter sensor. b) Our processing pipeline. Images travel directly from the camera to the graphics card (GPUDirect RDMA) to be recorded (GPUDirect Storage) and visualized (G-Sync) with minimal latency. Other sensors (e.g., lidar) can be added via the external FPGA-MIPI bridge. A network switch manages clock synchronization (PTP) across the vision pipeline and the surgical robot (dVRK). c) The components of the customized dVRK and its potential expansions to collaborative multi-console setups.}
    \label{Fig:Pipeline}
    \vspace{-4mm}
\end{figure*}

\section{Clinical Applications}
\label{Clinical}
Although telerobotic MIS was originally envisioned for long-distance surgery, upper bounds on communication speed and reliability have largely limited it to tethered local use. 
While improving ergonomics and reducing iatrogenic injuries compared to handheld MIS, single-camera surgical robots still constrain the field of view, preventing the surgeon from assessing anatomical relationships from other angles, as done in open surgery. This limitation is particularly relevant in complex procedures requiring delicate tissue handling or precise anatomical orientation, such as colorectal, urological, and gynecological surgeries, where deep pelvic visibility and depth perception are challenging from a single endoscopic perspective. Multiple viewpoints could reduce blind spots, improve spatial awareness, and therefore facilitate access to deep-seated or obscured structures, allowing advanced dissection and anastomosis.
    
Furthermore, the third robotic tool is mostly used for static tissue traction, since a surgical console can operate only two arms at a time. As a rare exception, dual-console setups allow a second surgeon to collaborate by moving the third instrument, which is otherwise frozen. Even in this setting, the maneuverability of the third tool depends on its orientation with respect to the only camera perspective shared by the two consoles, prohibiting some potentially interesting trocar configurations. Introducing a \textit{second active viewpoint} would facilitate such collaborative tasks by allowing the surgeons to adjust their views independently and maneuver the robotic instruments from different angles while maintaining intuitive wrist orientation. We foresee potential for such collaborative procedures, especially when one surgeon is remote. 
    
Ultimately, with increasing research on surgical autonomy for intraoperative tasks like camera control~\cite{DaCol21-FRAI-Usability}, suturing~\cite{kim2024surgical}, and diagnostics~\cite{Marahrens2022}, we believe the availability of \textit{synchronized multi-view 3D measurements} of the patient's anatomy could transformatively allow creation of robust and accurate deformable scene representations. Such advanced perception would not only facilitate multimodal imaging integration but would also enable the clinical translation of machine intelligence and shared autonomy. With accurate real-time intraoperative 3D perception, a single surgeon could receive assistance from an algorithmic assistant that directly controls one or more robotic instruments and/or robotic cameras.

\section{Proposed Technology}
\label{Technology} 

\subsection{Multi-endoscope Vision Pipeline}
When deploying a multi-camera system, especially to generate high-quality datasets for computer-vision research, \textit{hardware synchronization} is paramount. Current commercial stereo endoscopic towers do not expose their internal sync signal and cannot receive an external sync master. We therefore disassembled a high-quality clinical stereo endoscope and interfaced its optics with a custom sensing and processing pipeline. A large-format CMOS sensor captures both optical channels (Fig.~\ref{Fig:Pipeline}a) of each camera; its global shutter ensures perfect time alignment across all pixels. Frames travel via optical fibers directly from the camera's embedded FPGA to the computer's GPU where they can be processed (e.g., demosaicing, rectification, AI inference), sent to storage, and displayed on screen. The reference photon-to-glass latency (scene acquisition to monitor display) for this pipeline on a modern edge computer is 8\,ms for a 4K image. A network switch acts as a sync master sending Precision Time Protocol (PTP) signals across the local network that aligns ($<$\SI{1}{\micro\second}) the internal clocks of all connected devices (including the robot, Fig.~\ref{Fig:Pipeline}b). The stereo optics feature large depth of field (about 50--150\,mm) and fixed focus for stable camera calibration. Our prototypes could be deployed in vivo due to the scope's original disposable sleeve and draping.

\subsection{Multi-endoscope Surgical Robot}
To control two independent views, we modified both the hardware and the software of our da Vinci Research Kit (dVRK)~\cite{dettorre2021}. A second endoscopic camera manipulator (ECM) is mounted with a custom plate in place of the third patient side manipulator (PSM, Fig.~\ref{Fig:dVRK}). This swap was convenient to maintain the single-cart design of the robot, which features only four setup joint (SUJ) arms. Our minimal four-arm design could easily be extended by employing a different kinematic registration method for the additional arms. Furthermore, the SUJ design of the newer-generation dVRK-Si, identical across arms, will make such system expansions even more practical. We modified the dVRK software libraries to enable multiple ECMs in the control logic chain and to assign one or more PSMs to each ECM's respective kinematic reference frame. Finally, dVRK now also supports multiple surgeon consoles. These edits will allow users (human or algorithmic) to \textit{independently control multiple imaging devices} and \textit{simultaneously teleoperate instruments from different perspectives}.

\section{Impact and Future work}
\label{Impact}
We believe the proposed approach of multi-viewpoint surgical telerobotics will significantly benefit research and clinical development in computer-integrated surgery. By future release of our component specifications, mechanical designs, data-collection algorithms, and updates to the dVRK software libraries, we will provide the community with everything needed to reproduce our setup, improve it, and develop powerful algorithms for its successful deployment. While we present a minimal setup, the dVRK allows unlimited expansion, further fostering research creativity. Similarly, our vision pipeline -- built on industrial-grade components and standardized transfer protocols -- can be customized to support different imaging sensors and applications. We plan to conduct extensive data collection with the goal of producing high-quality synchronized multi-viewpoint datasets (including kinematics, stereo, and lidar) to boost the development of advanced computer-vision methods like neural scene representation and free-viewpoint photorealistic rendering. The aforementioned perception methods together with the proposed collaborative and semi-autonomous teleoperation paradigms will be developed with and thoroughly evaluated by clinicians to assess their feasibility and impact.\looseness-1

%%%%%%%%%%%%%%%%%%%%%%%%%%%%%%%%%%%%%%%%%%%%%%%%%%%%%%%%%%%%%%%%%%%%%%%%
\bibliographystyle{IEEEtran}
\bibliography{IEEEabrv, references_ICRA_RAMI_03_25}

\begin{thebibliography}{1}
\providecommand{\url}[1]{#1}
\csname url@samestyle\endcsname
\providecommand{\newblock}{\relax}
\providecommand{\bibinfo}[2]{#2}
\providecommand{\BIBentrySTDinterwordspacing}{\spaceskip=0pt\relax}
\providecommand{\BIBentryALTinterwordstretchfactor}{4}
\providecommand{\BIBentryALTinterwordspacing}{\spaceskip=\fontdimen2\font plus
\BIBentryALTinterwordstretchfactor\fontdimen3\font minus \fontdimen4\font\relax}
\providecommand{\BIBforeignlanguage}[2]{{%
\expandafter\ifx\csname l@#1\endcsname\relax
\typeout{** WARNING: IEEEtran.bst: No hyphenation pattern has been}%
\typeout{** loaded for the language `#1'. Using the pattern for}%
\typeout{** the default language instead.}%
\else
\language=\csname l@#1\endcsname
\fi
#2}}
\providecommand{\BIBdecl}{\relax}
\BIBdecl

\bibitem{Caccianiga2022DenseSurgery}
G.~Caccianiga and K.~J. Kuchenbecker, ``Dense {3D} reconstruction through lidar: A new perspective on computer-integrated surgery,'' in \emph{Proc. Hamlyn Symposium on Medical Robotics}, 2022, pp. 63--64.

\bibitem{simi2013}
M.~Simi, R.~Pickens, A.~Menciassi, S.~D. Herrell, and P.~Valdastri, ``Fine tilt tuning of a laparoscopic camera by local magnetic actuation: Two-port nephrectomy experience on human cadavers,'' \emph{Surgical Innovation}, vol.~20, no.~4, pp. 385--394, 2013, pMID: 23060534.

\bibitem{Avinash2019AStudy}
A.~Avinash, A.~E. Abdelaal, P.~Mathur, and S.~E. Salcudean, ``A ``pickup''' stereoscopic camera with visual-motor aligned control for the da {V}inci surgical system: a preliminary study,'' \emph{Int. Journal of Computer Assisted Radiology and Surgery}, vol.~14, no.~7, pp. 1197--1206, 2019.

\bibitem{DaCol21-FRAI-Usability}
T.~D. Col, G.~Caccianiga, M.~Catellani, A.~Mariani, M.~Ferro, G.~Cordima, E.~D. Momi, G.~Ferrigno, and O.~de~Cobelli, ``Automating endoscope motion in robotic surgery: A usability study on da {V}inci-assisted ex vivo neobladder reconstruction,'' \emph{Frontiers in Robotics and AI}, vol.~8, no. 707704, pp. 1--10, Nov. 2021.

\bibitem{kim2024surgical}
\BIBentryALTinterwordspacing
J.~W. Kim, T.~Z. Zhao, S.~Schmidgall, A.~Deguet, M.~Kobilarov, C.~Finn, and A.~Krieger, ``Surgical {R}obot {T}ransformer ({SRT}): Imitation learning for surgical tasks,'' in \emph{Conference on Robot Learning}, 2024.
\BIBentrySTDinterwordspacing

\bibitem{Marahrens2022}
\BIBentryALTinterwordspacing
N.~Marahrens, B.~Scaglioni, D.~Jones, R.~Prasad, C.~S. Biyani, and P.~Valdastri, ``Towards autonomous robotic minimally invasive ultrasound scanning and vessel reconstruction on non-planar surfaces,'' \emph{Frontiers in Robotics and AI}, vol.~9, 2022.
\BIBentrySTDinterwordspacing

\bibitem{dettorre2021}
C.~D’Ettorre, A.~Mariani, A.~Stilli, F.~Rodriguez~y Baena, P.~Valdastri, A.~Deguet, P.~Kazanzides, R.~H. Taylor, G.~S. Fischer, S.~P. DiMaio, A.~Menciassi, and D.~Stoyanov, ``Accelerating surgical robotics research: A review of 10 years with the da {V}inci {R}esearch {K}it,'' \emph{IEEE Robotics \& Automation Magazine}, vol.~28, no.~4, pp. 56--78, 2021.

\end{thebibliography}
\end{document}